\documentclass{article}

\usepackage{PRIMEarxiv}

\usepackage[utf8]{inputenc} 
\usepackage[T1]{fontenc}    
\usepackage{hyperref}       
\usepackage{url}            
\usepackage{booktabs}       
\usepackage{amsfonts}       
\usepackage{nicefrac}       
\usepackage{microtype}      
\usepackage{lipsum}
\usepackage{fancyhdr}       
\usepackage{graphicx}       
\graphicspath{{media/}}     
\usepackage{booktabs}
\usepackage{tabularx}
\usepackage[table]{xcolor}
\usepackage{enumitem}
\usepackage{caption}
\usepackage{subcaption}
\usepackage{hyperref}

\usepackage{amssymb}

\title{
A Survey of Agent Interoperability Protocols: Model Context Protocol (MCP), Agent Communication Protocol (ACP), Agent-to-Agent Protocol (A2A), and Agent Network Protocol (ANP)}

\author{
  Abul Ehtesham \\
  Kent State University \\
  Kent, OH, USA \\
  \texttt{aehtesha@kent.edu} \\
  \And
  Aditi Singh \\
  Cleveland State University \\
  Cleveland, OH, USA \\
  \texttt{a.singh22@csuohio.edu} \\
  \And
  Gaurav Kumar Gupta \\
  Youngstown State University \\
  Youngstown, OH, USA \\
  \texttt{gkgupta@student.ysu.edu} \\
  \And
  Saket Kumar \\
  Northeastern University \\
  Boston, MA, USA \\
  \texttt{kumar.sak@northeastern.edu} \\
}

\begin{document}
\maketitle

\begin{abstract}
Large language model (LLM)-powered autonomous agents demand robust, standardized protocols to integrate tools, share contextual data, and coordinate tasks across heterogeneous systems. Ad-hoc integrations are difficult to scale, secure, and generalize across domains. This survey examines four emerging agent communication protocols: Model Context Protocol (MCP), Agent Communication Protocol (ACP), Agent-to-Agent Protocol (A2A), and Agent Network Protocol (ANP), each addressing interoperability in distinct deployment contexts. MCP provides a JSON-RPC client-server interface for secure tool invocation and typed data exchange. ACP defines a general-purpose communication protocol over RESTful HTTP, supporting MIME-typed multipart messages and both synchronous and asynchronous interactions. Its lightweight and runtime-independent design enables scalable agent invocation, while features like session management, message routing, and integration with role-based and decentralized identifiers (DIDs).A2A enables peer-to-peer task delegation using capability-based Agent Cards, supporting secure and scalable collaboration across enterprise agent workflows. ANP supports open-network agent discovery and secure collaboration using W3C decentralized identifiers (DIDs) and JSON-LD graphs. The protocols are compared across multiple dimensions, including interaction modes, discovery mechanisms, communication patterns, and security models. Based on the comparative analysis, a phased adoption roadmap is proposed: beginning with MCP for tool access, followed by ACP for structured, multimodal messaging, session-aware interaction, and both online and offline agent discovery across scalable, HTTP-based deployments, A2A for collaborative task execution, and extending to ANP for decentralized agent marketplaces. This work provides a comprehensive foundation for designing secure, interoperable, and scalable ecosystems of LLM-powered agents.
\end{abstract}


\section{Introduction}

Large Language Models (LLMs) have become central to modern artificial intelligence, powering autonomous agents that operate across cloud, edge, and desktop environments~\cite{brown2020language,bommasani2021opportunities}. These agents~\cite{Wang_2024} ingest contextual information, execute tasks, and interact with external services or tools. However, inconsistent and fragmented interoperability practices make it difficult to integrate, secure, and scale communication among LLM-driven agents~\cite{mialon2023augmented}.

Interoperability (the ability of distinct agents and systems to discover capabilities, exchange context, and coordinate actions seamlessly) is essential for modular, reusable, and resilient multi-agent~\cite{guo2024largelanguagemodelbased} workflows. Standardized protocols reduce development overhead, improve security, and enable cross-platform collaboration. Clear, universally adopted standards remain nascent.

This survey examines four emerging agent communication protocols, each targeting a different interoperability tier:

\begin{itemize}
  \item \textbf{Model Context Protocol (MCP)}: A JSON-RPC client–server interface for secure context ingestion and structured tool invocation. MCP streamline the integration of large language models (LLMs) with external data sources and tools. MCP addresses the challenges of fragmented and custom-built integrations by providing a universal, model-agnostic interface for AI systems to access and interact with diverse resources~\cite{mcp2024introduction,singh2025mcp,ray2025mcp}. MCP was launched by Anthropic in Novemeber 2024.
  \item \textbf{Agent-to-Agent Protocol (A2A)}: Enable secure, structured, and interoperable collaboration between AI agents across platforms, vendors, and environments. A2A is designed to support peer-to-peer agent interactions using capability-based representations known as Agent Cards, which describe what an agent can do and how it can be securely invoked. A2A supports asynchronous, event-driven communication through HTTP and Server-Sent Events (SSE), making it suitable for distributed, scalable agent ecosystems,~\cite{google2024a2a}. A2A was launched by Google in April 2025.
  \item \textbf{Agent Communication Protocol (ACP)}: ACP is a general-purpose protocol for agent communication that uses RESTful HTTP interfaces to support MIME-typed multipart messages and both synchronous and asynchronous interactions. This REST-based communication model enables lightweight, runtime-independent agent invocation, making ACP well-suited for scalable system integration. ACP includes structured session management, message routing, and a flexible authentication model that integrates with role-based access control (RBAC) and decentralized identity (DID) systems. It is compatible with a wide range of agent frameworks and deployment models, from lightweight stateless utilities to long-running, stateful services. ACP also supports agent discovery through runtime APIs, offline packaging, and manifest-based metadata. ACP was launched by IBM in March 2025~\cite{beeai2024acp}.
  \item \textbf{Agent Network Protocol (ANP)}: The Agent Network Protocol (ANP) is an open-source communication framework designed to enable secure, decentralized collaboration among AI agents across the open internet. Unlike traditional client-server architectures, ANP adopts a peer-to-peer (P2P) model, allowing agents to autonomously discover, authenticate, and interact with one another without centralized intermediaries ~\cite{anp2024github, anp2024website}.
\end{itemize}

Architectural details, integration approaches, communication patterns, and security considerations are reviewed for each protocol. A comparison highlights trade-offs in interaction modes, discovery mechanisms, communication models, and security frameworks. A phased adoption roadmap sequences MCP, A2A, ACP, and ANP to guide progressive deployment in real-world agent ecosystems.

The remainder of the paper is organized as follows. Section 2 discusses challenges in agent interoperability. Section 3 reviews background and related work. Sections 4–7 describe the architectures of MCP, A2A, ACP, and ANP, respectively. Section 8 presents the comparative evaluation. Section 9 outlines the phased adoption roadmap. Section 10 concludes and suggests future research directions.




\section{Challenges and Solutions in Agent Protocol Interoperability}
Despite the emergence of multiple open protocols like MCP, ACP, A2A, and ANP, achieving seamless agent interoperability in real-world AI systems remains a non-trivial task. This section identifies key challenges encountered in agent-based architectures and highlights how each protocol addresses them with purpose-built design principles.

\textbf{Lack of Context Standardization for LLMs:}
Large Language Models (LLMs) require contextual grounding to produce accurate outputs. However, existing application architectures provide no unified mechanism to deliver structured context to LLMs, leading to ad hoc tool integrations and unreliable behavior.
\textit{Solution:} The Model Context Protocol (MCP) addresses this by standardizing how applications deliver tools, datasets, and sampling instructions to LLMs, akin to a USB-C for AI. It supports flexible plug-and-play tools, safe infrastructure integration, and compatibility across LLM vendors.

\begin{figure}[b]
    \centering
    \includegraphics[width=0.5\linewidth]{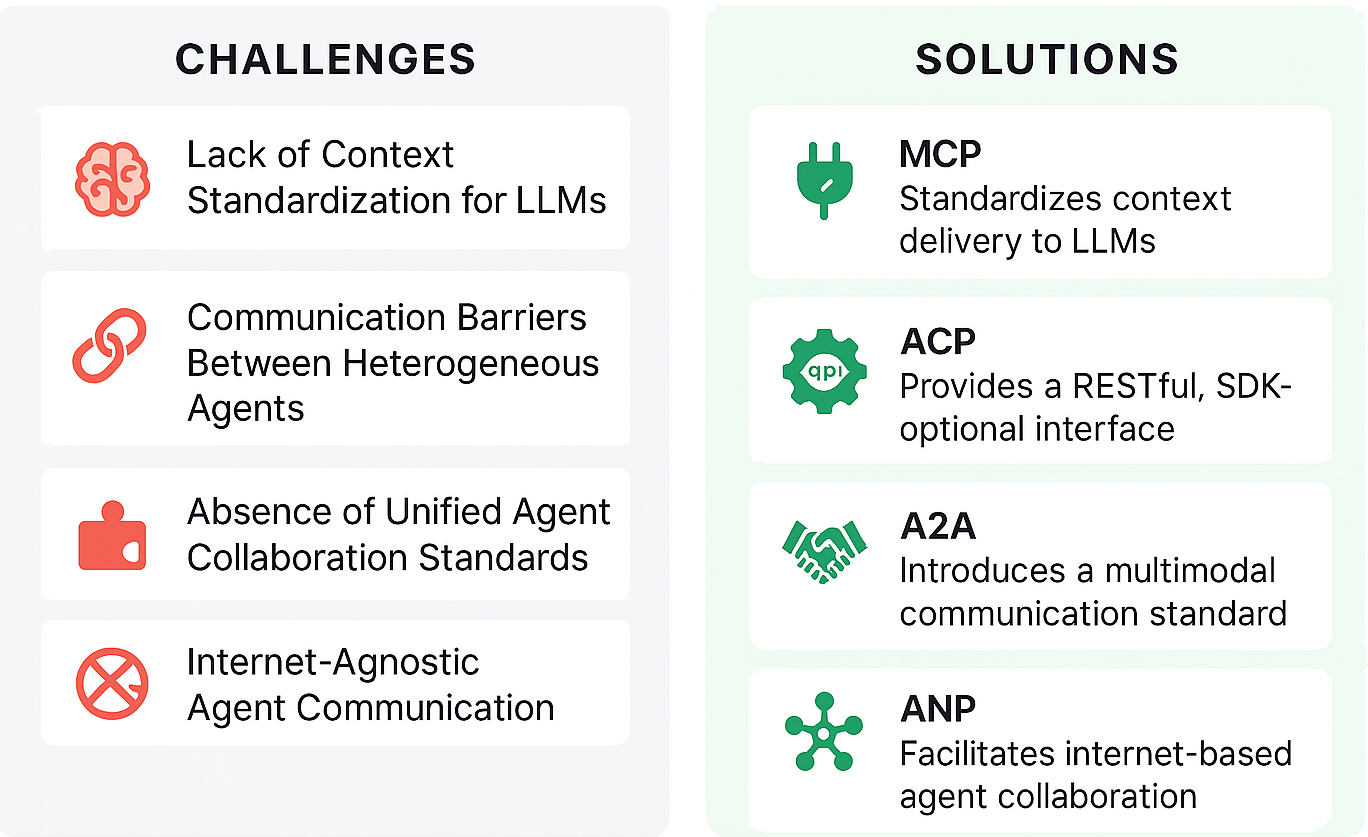}
    \caption{Protocol-aligned solution to challenges in agent communication}
    \label{fig:enter-label}
\end{figure}

\textbf{Communication Barriers Between Heterogeneous Agents:}
Enterprise systems often consist of agents built using different stacks and frameworks, resulting in isolated behavior and poor collaboration.
\textit{Solution:} The Agent Communication Protocol (ACP) offers a RESTful, SDK-optional interface with open governance under the Linux Foundation. It enables asynchronous-first interactions, offline discovery, and vendor-neutral execution, bridging interoperability gaps at scale.

\textbf{Absence of Unified Agent Collaboration Standards:}
Even when agents communicate, there's no shared framework for dynamic negotiation, capability sharing, and coordination.
\textit{Solution:} The Agent2Agent (A2A) protocol introduces a multimodal communication standard to unlock dynamic interaction between opaque, autonomous agents—regardless of framework. It simplifies enterprise integration and supports shared task management and user experience negotiation.

\textbf{Internet-Agnostic Agent Communication:}
The modern internet is optimized for human interaction but suboptimal for autonomous agents, which require low-latency, API-native communication and decentralized identity validation.
\textit{Solution:} The Agent Network Protocol (ANP) provides a layered protocol architecture incorporating decentralized identity (W3C DID), semantic web principles, and encrypted communication to facilitate cross-platform agent collaboration over the open internet.

Together, these protocols aim to transform fragmented AI ecosystems into robust, secure, and interoperable agent networks scalable across organizational and vendor boundaries. See Table~\ref{tab:agent-protocol-comparison} for a detailed comparative overview.





\section{Background and Related Work}
\label{sec:background}

Autonomous agents powered by large language models (LLMs) are rapidly being adopted across industries to automate complex tasks, yet disparate frameworks and ad-hoc integrations hinder robust interoperability, security, and scalability~\cite{bommasani2021opportunities,mialon2023augmented}.  Recent surveys have begun to characterize the landscape of LLM-based multi-agent systems, categorizing collaboration patterns, memory architectures, and orchestration strategies~\cite{tran2025multi,guo2024survey,yan2025beyond}.  However, these works largely focus on high-level workflows and neglect the underlying protocols necessary for dynamic peer discovery, capability negotiation, and secure tool invocation.

Effective interoperability—enabling agents to discover capabilities, share context, and coordinate actions—is critical for building modular, reusable, and resilient multi-agent systems.  Early efforts in dynamic discovery have introduced metadata manifests and capability descriptors to allow runtime agent registration and lookup~\cite{sheriff2024dynamic}, while recent work on automated tool testing frameworks (e.g., TOOLFUZZ) highlights the challenges of ensuring compatibility across evolving API surfaces~\cite{milev2024toolfuzz}.  Yet, no unified protocol has emerged that specifies how agents should announce their interfaces, authenticate peers, or negotiate context sharing across heterogeneous LLM frameworks.

In response to this gap, recent proposals such as the Model Context Protocol (MCP), Agent Communication Protocol (ACP), Agent2Agent Protocol (A2A)  and Agent Network Protocol (ANP) aim to define lightweight, formal interfaces for context ingestion, performative messaging, and peer discovery using JSON-RPC schemas~\cite{mcp2024introduction,beeai2024acp,google2024a2a, anp2024github, anp2024website}. Each protocol is examined in detail, followed by a comparative analysis and a roadmap for their integration within emerging multi-agent ecosystems.

\subsection{AI Agents: Definition and Scope}
\label{sec:definition}

An \emph{AI agent} is defined as any autonomous software entity that perceives its environment through inputs (e.g., user queries, sensor data) and acts upon it via outputs (e.g., API calls, messages) to achieve designated goals~\cite{russell2010aima}.  Agents operate within environments characterized along dimensions such as observability, determinism, episodicity, and dynamicity, and may employ sensors and actuators to interact with physical or virtual world models~\cite{russell2010aima,wooldridge2009introduction}.  

According to Franklin and Graesser’s taxonomy, agents can be categorized based on attributes like autonomy, sociability, reactivity, and adaptability, reflecting their ability to function in open, multi-agent settings~\cite{franklin1997taxonomy}.  Jennings emphasizes proactive goal generation, complex planning, and robust recovery capabilities under uncertainty as key distinguishing features from simple reactive programs~\cite{jennings2000agent}.  Wooldridge further identifies four core properties—\emph{autonomy}, \emph{social ability}, \emph{reactivity}, and \emph{pro‐activeness}, that enables agents to operate without direct human intervention, collaborate with peers, and pursue long‐term objectives~\cite{wooldridge2009introduction}.  

Agent architectures span from simple rule‐based reactive models, where actions are direct responses to percepts, to rich deliberative frameworks such as Belief‐Desire‐Intention (BDI) systems that support symbolic reasoning, dynamic plan execution, and intention reconsideration.  In multi-agent systems, coordination is achieved through communication protocols, negotiation strategies, and organizational structures, laying the groundwork for LLM‐powered ecosystems that require robust interoperability, security, and scalability.  This broad yet precise definition underpins our subsequent review of communication standards, orchestration frameworks, and protocol designs.

\subsection{Early Symbolic Agent Languages—Evolution of Agent Communication Standards}
The first formal agent messaging languages emerged in the early 1990s with the goal of providing a standardized “envelope” and performative vocabulary for knowledge‐based systems.  The \textbf{Knowledge Query and Manipulation Language (KQML)} introduced by Genesereth and Ketchpel defined a set of speech‐act performatives (e.g., \texttt{ask‐if}, \texttt{tell}, \texttt{reply}) along with a flexible message envelope supporting parameters such as \texttt{:content}, \texttt{:language}, \texttt{:ontology}, \texttt{:receiver}, and \texttt{:reply‐with}.  KQML also specified content‐language bindings (commonly KIF) to express propositions in a machine‐interpretable form~\cite{genesereth1993kqml,finin1994kqml}.  Although widely used in DARPA’s Open Knowledge Base and Agent projects, KQML’s lack of formal semantics for performatives and heavyweight XML‐style encodings hindered large‐scale deployments.

Building on KQML, the \textbf{FIPA Agent Communication Language (FIPA‐ACL)}—ratified by the Foundation for Intelligent Physical Agents in 2000—refined the notion of communicative acts by prescribing precise pre‐ and post‐condition semantics grounded in agents’ mental states (beliefs, desires, intentions).  FIPA‐ACL defined a richer set of performatives (e.g., \texttt{agree}, \texttt{refuse}, \texttt{request}), standardized content languages (e.g., SL0, SL1), and outlined interaction protocols for common patterns such as \emph{contract net}, \emph{iterated contract net}, and \emph{subscribe/notify}~\cite{fipa2000acl}.  Reference implementations in platforms like JADE and JACK offered Java‐based agent containers and message‐handling APIs, yet the complexity of FIPA’s ontology management, coupled with verbose XML encodings, limited its uptake to academic and defense use cases rather than lightweight, industry‐grade systems.

\subsection{Service-Oriented Integrations and Retrieval-Augmented Generation}
The early 2000s witnessed the rise of service-oriented architectures (SOA), in which enterprise systems exposed functionality as web services (SOAP, WSDL, WS-* standards) and registered endpoints in UDDI repositories~\cite{curbera2002web}. Message-oriented middleware and enterprise service buses (ESBs) such as Apache Camel and Mule ESB facilitated protocol bridging, message routing, and payload transformation, leveraging patterns like content-based routing, message splitting, and aggregation~\cite{hohpe2006enterprise}. While SOA and ESBs decoupled service producers from consumers, they often incurred high operational complexity, brittle adapters, and configuration sprawl as APIs evolved and security requirements tightened.

With the advent of large language models, \textbf{Retrieval-Augmented Generation (RAG)} emerged in 2020 to integrate external knowledge into generation pipelines by coupling dense vector retrieval with autoregressive decoding~\cite{lewis2020rag}. RAG systems encode queries and documents in a shared embedding space (e.g., DPR) to fetch top-$k$ relevant passages, then condition LLM outputs on retrieved context to reduce hallucinations and enable dynamic knowledge updates~\cite{izacard2021towards}. Despite improving factuality and flexibility, RAG frameworks treat retrieval and generation as separate batch processes and do not prescribe how LLMs should translate grounded content into executable actions or orchestrate multi-step workflows—highlighting a need for protocol-level standards that unify knowledge grounding with action invocation.

\subsection{LLM Agents and Function Calling}
The rapid evolution of large language models (LLMs) such as GPT-3.5, GPT-4, Claude, and Gemini has fundamentally transformed agent design by enabling zero- and few-shot understanding of complex natural language instructions without bespoke rule engines~\cite{bommasani2021opportunities}.  These foundation models can parse user intent, plan multi-step workflows, and maintain dialogue coherence across diverse domains, opening the door to “LLM agents” that combine linguistic reasoning with external tool execution.

To operationalize tool use, OpenAI introduced \textbf{function calling} in 2023, a lightweight protocol whereby an LLM can output a JSON-formatted signature corresponding to a predefined API endpoint~\cite{openai2023function}.  Under this paradigm, developers supply the model with a catalog of function definitions—each described by a name, JSON schema for arguments, and descriptive help text—and the model decides at generation time whether to invoke a function, emitting well-formed JSON that can be parsed and executed by downstream systems.  This approach unifies natural language understanding and action invocation, enabling real-time data fetches, database queries, and transactional operations from within a single LLM response.

Building on this core capability, several frameworks have emerged to simplify agent development:
\begin{itemize}
  \item \textbf{LangChain} provides abstractions for chaining LLM calls, memory buffers, and function invocation in modular workflows, with built-in support for retrievers, vector stores, and agent loops~\cite{chase2022langchain}.
  \item \textbf{LlamaIndex} (formerly GPT Index) focuses on integrating LLMs with custom knowledge bases, offering document loaders, index wrappers, and a “tool registry” that maps user queries to API calls~\cite{wu2023llamaindex}.
  \item The \textbf{OpenAI Plugin Store} enables third-party tool providers to register plugins that expose RESTful interfaces, metadata, and authentication flows, which can be discovered and invoked by any model with plugin access~\cite{openai2023plugin}.
\end{itemize}

Despite these advances, current function-calling ecosystems suffer from several limitations.  Tool definitions are typically static: agents must be re-initialized whenever new APIs are added or schemas change, preventing truly dynamic discovery.  Security boundaries—such as authentication tokens, rate limits, and access control—are ad-hoc and framework-specific, increasing the risk of unauthorized calls.  Moreover, each framework employs its own metadata conventions, hindering cross-framework reuse of tools and requiring bespoke adapters for interoperability~\cite{liu2024autotool}.  Addressing these challenges requires protocol-level standards that prescribe a common schema for function metadata, dynamic capability negotiation, and end-to-end security guarantees across heterogeneous LLM agent platforms.

\subsection{Orchestration and Lightweight Agent Frameworks}
Recent advances have extended the capabilities of LLMs beyond reasoning to include orchestration of  external tool invocation. \textbf{Toolformer} employs a self-supervised masking strategy that exposes potential API calls during pretraining, enabling the model to learn when and how to invoke functions as part of its text generation~\cite{le2023toolformer}.  \textbf{ReAct} interleaves chain-of-thought reasoning with explicit action calls, allowing models to alternate between “thinking” steps and tool invocations based on intermediate observations~\cite{yao2023react}.  These approaches unify reasoning and action at the single-agent level but do not address peer discovery or multi-agent coordination.

Complementing these algorithmic techniques, several lightweight frameworks have emerged to suppprt multi-agent orchestration with minimal boilerplate (Table~\ref{tab:frameworks}). Additional orchestration systems, such as AutoGPT’s autonomous loops~\cite{autogen2023} and Reflexion’s iterative self-improvement mechanism~\cite{shinn2023reflexion}, highlight the value of feedback and adaptation in agent workflows, However, these frameworks continue to rely on static tool registries and bespoke communication layers.  Across these approaches, the lack of a standardized protocol for capability advertisement, peer authentication, and cross-framework composition contributes to fragmentation-hindering the emergence of a cohesive, interoperable agent ecosystem.

\begin{table}[!t]
  \centering
  \renewcommand{\arraystretch}{1.3}
  \rowcolors{2}{gray!10}{white}
  \captionsetup{justification=centering}
  \caption{Lightweight LLM Agent Frameworks}
  \label{tab:frameworks}
  \begin{tabularx}{\textwidth}{|
    >{\centering\arraybackslash}p{2.5cm}|
    >{\raggedright\arraybackslash}X|
    >{\raggedright\arraybackslash}p{2.5cm}|}
    \hline
    \rowcolor{gray!30}
    \textbf{Framework} & \textbf{Core Feature} & \textbf{Reference} \\
    \hline
    CrewAI            & High-level crew abstractions for role assignment, subtask delegation, and message routing among agents      & \cite{crewai2024}       \\
    SmolAgents        & Single-file Python library combining retrieval, vision, and agent loop primitives for rapid prototyping           & \cite{smolagents2024}   \\
    AG2 (AutoGen)     & Open-source AgentOS with human-in-the-loop checkpoints, policy enforcement hooks, and lifecycle management       & \cite{ag22025}          \\
    Semantic Kernel   & Enterprise-grade SDK unifying memory stores, planning modules, and plugin orchestration across sessions        & \cite{semantickernel2024}\\
    Swarm             & Stateless multi-agent coordination via JSON-RPC routines, spawning and aggregating parallel agent tasks         & \cite{openai2024swarm}  \\
    \bottomrule
  \end{tabularx}
\end{table}

\subsection{Protocol Evolution Timeline}
\label{sec:timeline}

\begin{figure}
    \centering
    \includegraphics[width=0.4\linewidth]{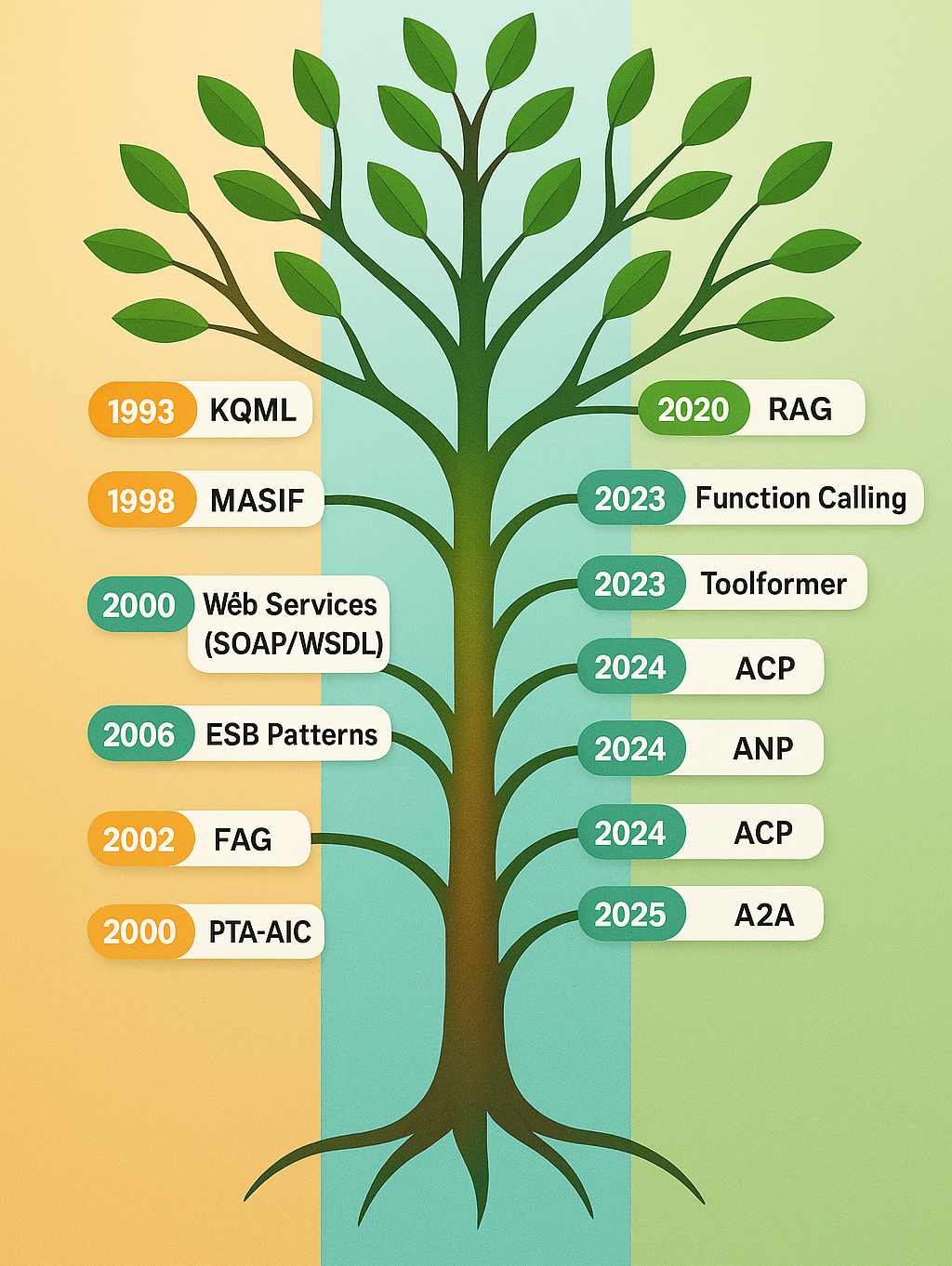}
    \caption{Timeline of Interoperability}
    \label{fig:timeline}
\end{figure}

The evolution of agent interoperability is illustrated through a visual timeline (Figure~\ref{fig:timeline}) and a detailed table (Table~\ref{tab:timeline}). The timeline captures high-level milestones, while the table offers technical detail, describing each development alongside key contributions. Together, these representations outline the trajectory of interoperability standards and protocols over time.



\begin{table}[!t]
\small
  \centering
  \renewcommand{\arraystretch}{1.3}
  \rowcolors{2}{gray!10}{white}
  \captionsetup{justification=centering}
  \caption{Timeline of Key Agent Interoperability Milestones}
  \label{tab:timeline}
  
  \begin{tabularx}{\textwidth}{|
    >{\centering\arraybackslash}p{1.5cm}|
    >{\raggedright\arraybackslash}p{2.5cm}|
    >{\raggedright\arraybackslash}X|}
    \hline
    \rowcolor{gray!30}
    \textbf{Year} & \textbf{Milestone} & \textbf{Key Contribution} \\
     \hline
    1993 & KQML & Introduced speech-act primitives and a flexible message envelope for knowledge-based agents~\cite{genesereth1993kqml}. \\
    1998 & MASIF & Defined basic service registration and discovery mechanisms for agent environments~\cite{pitt1998masif}. \\
    2000 & FIPA-ACL & Standardized performative semantics, content languages, and interaction protocols with formal pre-/post-conditions~\cite{fipa2000acl}. \\
    2002 & Web Services (SOAP/WSDL) & Enabled service-oriented agent integration via UDDI, XML messaging, and contract definitions~\cite{curbera2002web}. \\
    2006 & ESB Patterns & Codified enterprise integration patterns (routing, transformation) in ESBs like Apache Camel and Mule~\cite{hohpe2006enterprise}. \\
    2020 & RAG & Coupled dense vector retrieval with LLM decoding to ground outputs in external corpora~\cite{lewis2020rag}. \\
    2023 & Function Calling & Allowed LLMs to emit JSON-formatted API calls against a catalog of function schemas~\cite{openai2023function}. \\
    2023 & Toolformer & Trained LLMs via self-supervised masking to predict API call placement in text~\cite{le2023toolformer}. \\
    2023 & ReAct & Interleaved chain-of-thought reasoning and explicit action calls for dynamic workflows~\cite{yao2023react}. \\
    2024 & MCP & Proposed a JSON-RPC protocol for standardized context ingestion and tool invocation~\cite{mcp2024introduction}. \\
     2024 & ANP & Peer-to-peer protocol enabling cross-platform and cross-organization agent communication over the open internet.~\cite{anp2024github}. \\
    2024 & ACP & Defined performative messaging primitives with formal types and security layers~\cite{beeai2024acp}. \\
    2025 & A2A & Introduced peer discovery, capability exchange, and decentralized agent dialogues~\cite{google2024a2a}. \\
    \bottomrule
  \end{tabularx}
\end{table}

Three distinct evolutionary phases emerge:

\begin{enumerate}
    \item \textbf{Symbolic and SOA Foundations (1993–2006):} Early interoperability standards such as KQML and FIPA-ACL set formal semantic foundations. Subsequent developments in Web Services and Enterprise Service Bus (ESB) frameworks streamlined enterprise integration but introduced complexity and limited flexibility.
    
    \item \textbf{Retrieval and In-Model Action (2020–2023):} Marked by the introduction of Retrieval-Augmented Generation (RAG), this phase leveraged vector-based retrieval to enhance the grounding of language model outputs. Innovations like Function Calling, Toolformer, and ReAct enabled LLMs to directly translate reasoning into executable API calls, significantly advancing agent autonomy and flexibility.
    
    \item \textbf{Protocol-Oriented Interoperability (2024–2025):} The current phase emphasizes lightweight, standardized protocols such as MCP, ACP, ANP, and A2A. These protocols address previous limitations by enabling dynamic discovery, secure communication, and decentralized collaboration across heterogeneous agent systems, promoting scalability and robust interoperability.
\end{enumerate}

\section{MCP}

\subsection{Client Application (Host)}
The \textbf{Client Application (Host)} serves as the initiator of interactions in the MCP ecosystem. It is responsible for managing connections to one or more MCP Servers and orchestrating communication workflows in accordance with protocol specifications. In practice, the client initializes sessions, requests and processes the four core primitives \textit{Resources}, \textit{Tools}, \textit{Prompts}, and \textit{Sampling}, and handles asynchronous notifications related to server-side events. The client must also implement robust error-handling routines to gracefully manage communication failures or timeout conditions, ensuring reliable coordination with remote MCP Servers.


\begin{figure}
    \centering
    \includegraphics[width=\linewidth]{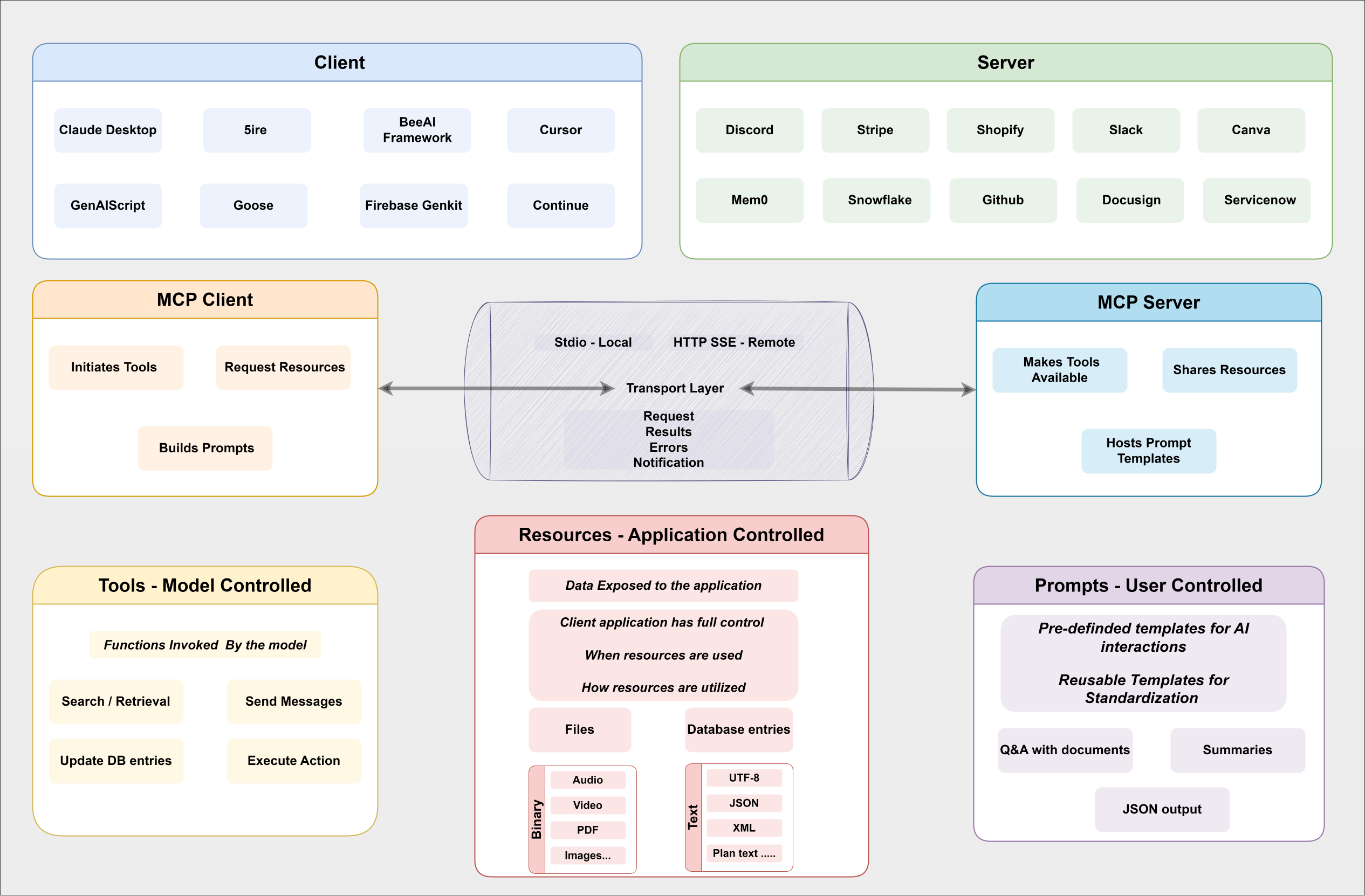}
    \caption{An overview of MCP~\cite{singh2025mcp}}
    \label{fig:MCP}
\end{figure}

\subsection{MCP Server (Providing Context \& Capabilities)}
The \textbf{MCP Server} functions as the provider of data, services, and interaction templates that the client can utilize to enrich LLM-based workflows. It exposes and manages contextual \textit{Resources}, executes external operations via \textit{Tools}, defines reusable \textit{Prompts} for consistent interaction patterns, and optionally delegates text-generation tasks through \textit{Sampling}. Beyond serving requests, the server is responsible for enforcing access control policies, maintaining operational security, and emitting notifications that reflect changes in its available capabilities. This provider-side architecture complements the client’s orchestration logic by modularizing access to complex or dynamic resources.

\subsection{Core Components}
The Model Context Protocol is composed of several layered abstractions that govern the structure and semantics of communication. At the foundation lies the \textbf{Protocol Layer}, which defines the semantics of message exchange using the JSON-RPC 2.0 specification. It ensures that each request is linked to a corresponding response and that all interactions conform to predictable patterns. Above this, the \textbf{Transport Layer} handles the physical transmission of messages between the client and server, supporting both local communication via \texttt{Stdio} and network-based channels such as \texttt{HTTP} with optional Server-Sent Events (SSE). At the highest abstraction, MCP organizes messages into four types: \textbf{Requests}, which are calls expecting replies; \textbf{Results}, which are successful responses to earlier requests; and \textbf{Errors}, which indicate failures or invalid invocations. A fourth type, \textbf{Notifications}, is used for asynchronous updates that do not require a client acknowledgment.

\subsection{MCP Server Core Capabilities}

The MCP Server offers four core capabilities \textbf{Tools}, \textbf{Resources}, \textbf{Prompts}, and \textbf{Sampling} each mapped to a distinct control model that governs the interaction between the client, the server, and the LLM. 

\textbf{Tools} are model-controlled capabilities that allow the LLM to invoke external APIs or services, often automatically and sometimes with user approval. This facilitates seamless integration with third-party systems and streamlines access to real-world data and operations. 

\textbf{Resources} are application-controlled elements, such as structured documents or contextual datasets, that are selected and managed by the client application. They provide the LLM with tailored, task-specific inputs and enable context-aware completions. 

\textbf{Prompts} are user-controlled templates defined by the server but selected by end-users through the client interface. These reusable prompts promote consistency, reduce redundancy, and support repeatable interaction patterns.

\textbf{Sampling} is server-controlled and allows the MCP Server to delegate the task of generating LLM completions to the client. This supports sophisticated agentic workflows and enables fine-grained oversight over the model’s generative process, including the ability to adjust temperature, length, and other sampling parameters dynamically.

\subsection{MCP Connection Lifecycle}

The Model Context Protocol (MCP) defines a three-phase lifecycle for client–server interactions, designed to ensure robust session management, secure capability negotiation, and clean termination. These phases \textbf{Initialization}, \textbf{Operation}, and \textbf{Shutdown} correspond to the temporal sequence of communication between the Client Application and MCP Server.

\textbf{Initialization} begins by establishing protocol compatibility and exchanging supported capabilities. During version negotiation, the client and server agree on the highest mutually supported protocol version. This is followed by a capability exchange, in which both sides advertise optional features—such as sampling, prompts, tools, and logging—that can be used during the session. The phase concludes with a \texttt{notifications/initialized} message sent by the client after receiving the server’s \texttt{initialize} response, signaling readiness to proceed to operational communication.

\textbf{Operation} represents the core active phase, during which the client and server exchange JSON-RPC method calls and notifications in accordance with the negotiated capabilities. Both parties are expected to adhere strictly to the features agreed upon during initialization, ensuring compatibility and predictability. Each task invocation may include a configurable timeout, and if a response is not received within that window, the client may issue a cancellation notification to prevent resource exhaustion or stale execution threads.

\textbf{Shutdown} ensures a clean and predictable end to the session. Either party may initiate termination by closing the transport layer typically HTTP or stdio which signals the end of communication. Upon shutdown, both client and server are responsible for resource cleanup, including the removal of active timeouts, cancellation of subscriptions, and deallocation of any spawned child processes. After this point, no new protocol messages should be sent, with the exception of essential diagnostics like ping or log flush events.

\subsection{Security Challenges and Mitigations Across the MCP Lifecycle}
As MCP adoption increases in enterprise and developer ecosystems, its lifecycle introduces multiple security vulnerabilities spanning initialization, operation, and update phases. These risks include tool poisoning, privilege persistence, and command injection, among others, many of which are amplified by LLMs’ susceptibility to prompt manipulation and opaque execution traces. 

Table~\ref{tab:mcp-threats} summarizes the most critical security threats identified across each lifecycle phase of MCP deployments, alongside their corresponding mitigation strategies and authoritative references. This synthesis reflects both current attack disclosures and best-practice defenses from recent audits and protocol reviews.

\begin{table}[ht]
  \centering
  \renewcommand{\arraystretch}{1.2}
  \rowcolors{2}{gray!10}{white}
  \captionsetup{justification=centering}
  \caption{Threats and Mitigation Strategies Across the MCP Lifecycle}
  \label{tab:mcp-threats}
\begin{tabularx}{\textwidth}{|
    >{\centering\arraybackslash}p{2.1cm}|
    >{\raggedright\arraybackslash}X|
    >{\raggedright\arraybackslash}X|
    >{\raggedright\arraybackslash}X|}

    \hline
    \rowcolor{gray!30}
    \textbf{Phase} & \textbf{Threat} & \textbf{Description} & \textbf{Mitigation Strategy} \\
    \hline

    \textbf{Creation} 
    & Installer Spoofing 
    & Malicious packages introduced during build or install pipelines.
    & Enforce SBOMs, digital signatures, and reproducible builds. \\

    & Supply-Chain Backdoors 
    & Persistent malware via CI/CD artifacts.
    & Harden CI/CD, validate manifests, and verify artifact integrity. \\

    & Name Collision 
    & Impersonation of trusted MCP agents using similar names.
    & Use Sigstore and DIDs to ensure unique, verifiable identities. \\

    & No Auth Handshake 
    & Clients connect to unauthenticated or rogue servers.
    & Enforce mutual authentication and TLS-based validation. \\

    \hline

    \textbf{Operation} 
    & Tool Poisoning 
    & Malicious prompts or metadata influencing LLM behavior.
    & Validate schemas, use filtering (YARA/RegEx), and apply semantic guards. \\

    & Credential Theft 
    & Secrets leaked via completions or tool output.
    & Use OAuth 2.1 + PKCE, restrict token scopes, and enforce mTLS. \\

    & Sandbox Escape 
    & Tools access host OS or bypass isolation.
    & Use syscall filters, AppArmor, and container hardening. \\

    & Remote Access Control 
    & LLMs inject SSH keys or create backdoor shells.
    & Monitor with EDR/HIDS and restrict outbound behavior. \\

    & Command Injection / RCE 
    & Unsafe inputs trigger system execution.
    & Sanitize inputs, disable shell access, and disallow evals. \\

    & Tool Redefinition 
    & Tools turn malicious after validation ("rug pull").
    & Use signed, versioned manifests and monitor for mutation. \\

    & Cross-Server Shadowing 
    & One server overrides another’s tool references.
    & Enforce scoped namespaces and validate routing origins. \\

    & Lack of Visibility 
    & Clients cannot inspect tool instructions or payloads.
    & Enable debug mode, metadata introspection, and logging. \\

    \hline

    \textbf{Update} 
    & Version Drift 
    & Older vulnerable MCP versions remain in use.
    & Use GitOps for drift detection and enforce auto-remediation. \\

    & Privilege Persistence 
    & Retained elevated roles or old token scopes.
    & Audit roles after updates and rotate credentials. \\

    & Configuration Drift 
    & Misconfigurations introduced post-update.
    & Validate against CVEs and apply hardened defaults. \\

    & Unsigned Tool Manifests 
    & Manifests altered or injected post-deployment.
    & Enforce signature checks and block unsigned tools. \\

    \hline
  \end{tabularx}
\end{table}

\section{A2A Architecture}
The Agent-to-Agent (A2A) architecture facilitates communication and collaboration between distinct agentic systems to accomplish tasks. It comprises three primary actors—\textit{User}, \textit{Client Agent}, and \textit{Remote Agent (Server)}, that interact via a well-defined protocol, enabling secure and interoperable execution.

\begin{figure}
    \centering
    \includegraphics[width=\linewidth]{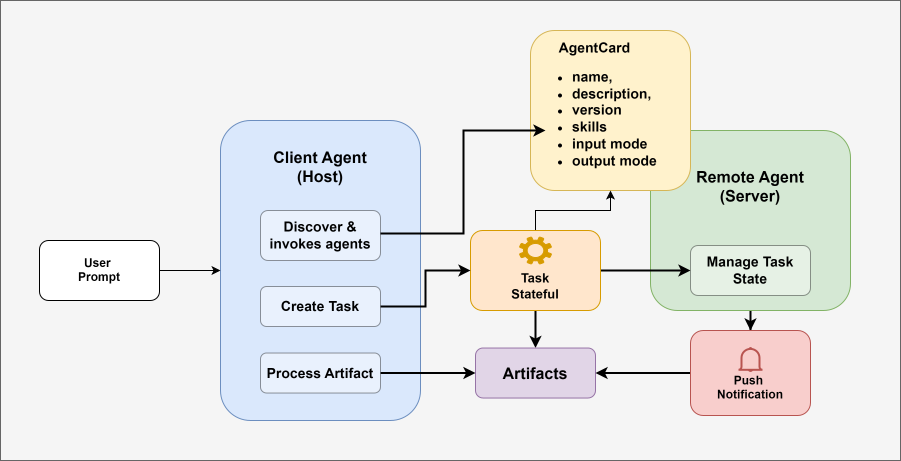}
    \caption{An overview of A2A}
    \label{fig:A2A}
\end{figure}

\subsection{Core Components}
The \textbf{User} initiates a task or request, typically without needing to understand or directly interact with the underlying agentic systems. The \textbf{Client Agent} receives this request, analyzes its intent, and identifies a suitable \textbf{Remote Agent (Server)} by inspecting the advertised capabilities through its Agent Card. Once selected, the Client Agent engages the Remote Agent to execute the task, coordinating message exchanges and retrieving results—termed \textit{Artifacts}—which are then delivered back to the User.

\subsubsection{User}
The \textbf{User} acts as the originator of any A2A interaction, embodying the intent or need that sets the agentic process in motion. While frequently a human end-user, the User can also be a system, service, or another agent in hierarchical workflows. Regardless of its form, the User does not directly interact with Remote Agents; instead, it relies on the Client Agent to translate its requests into actionable tasks and to mediate all responses.

A2A supports diverse User interaction models. A \textbf{Direct End-User} may engage with the Client Agent through interfaces such as chatbots or voice assistants, providing task input and receiving results in real time. An \textbf{Indirect End-User} interacts with higher-level systems that transparently utilize A2A agents behind the scenes, such as enterprise dashboards or orchestration tools. \textbf{Systems or Services} may also act as Users, invoking A2A agents autonomously for workflows like data transformation or monitoring. In multi-agent hierarchies, an \textbf{Agent as User} scenario occurs when one agent triggers downstream actions by another agent to fulfill complex tasks. These various User paradigms underscore the protocol’s agnosticism toward User identity and emphasize its focus on standardizing communication between Client and Remote Agents.

\subsubsection{Client Agent}
The \textbf{Client Agent} serves as an intermediary that represents the User's intent and coordinates with Remote Agents to fulfill it. Its responsibilities span multiple stages of the task lifecycle. It begins by performing \textbf{Agent Discovery}, retrieving and evaluating Agent Cards that describe each Remote Agent’s skills, capabilities, input/output specifications, and authentication requirements. Based on this discovery, the Client selects a Remote Agent aligned with the User’s task. Next, the Client Agent is responsible for \textbf{Task Initiation}. It constructs a structured Task object, encapsulating the User's intent, relevant metadata, and formatted inputs. It then sends this task to the selected Remote Agent using a well-formed Message. During execution, the Client Agent manages the \textbf{Message and Artifact Exchange}. It communicates bi-directionally with the Remote Agent, sending new instructions or follow-ups, and receiving outputs—termed \textit{Artifacts}—along with any intermediate updates. For long-running or stateful interactions, it maintains \textbf{Session Context}, using identifiers to group related exchanges under a unified workflow.

The Client Agent also oversees \textbf{Error Handling}, parsing any failure responses returned by the Remote Agent, and executing appropriate recovery strategies such as retries, fallback agent selection, or User notifications. After execution, the \textbf{Result Presentation} step involves transforming Artifacts into a user-consumable format and integrating them into the surrounding application or user interface.

Where supported, the Client Agent may handle \textbf{Asynchronous Communication} through mechanisms such as Server-Sent Events (SSE) or push notifications. For SSE, it establishes a persistent connection and streams updates in real time. If push notification support is available, the Client Agent registers with a notification service to receive task updates delivered out-of-band. Altogether, the Client Agent acts as the execution orchestrator, data translator, and communication bridge within the A2A protocol, enabling intelligent, context-aware interactions on behalf of the User.

\subsubsection{Remote Agent (Server)}
The \textbf{Remote Agent (Server)} is the service endpoint that executes tasks delegated by the Client Agent. It provides one or more \textbf{Skills}, which represent discrete operations it can perform ranging from simple data retrieval to complex computations or orchestrations involving external APIs or databases. Each skill is formally defined by its input and output schema, enabling consistent invocation across clients.

To make these capabilities discoverable, the Remote Agent publishes an \textbf{Agent Card}—a structured metadata document that includes a list of available skills, usage instructions, input/output formats, supported protocols, and authentication requirements. This Agent Card acts as both an advertisement and an interface contract for interacting agents.

The Remote Agent must also manage its internal \textbf{Resource Usage}, ensuring fair allocation of compute, memory, network, and storage resources during task execution. Alongside execution, it is responsible for enforcing \textbf{Security and Access Control} mechanisms. This includes authenticating Clients, verifying message integrity, and authorizing access to specific skills based on access policies or token scopes. By abstracting service capabilities into modular, independently managed components, the Remote Agent supports composability, reliability, and interoperability within agentic ecosystems.

\subsection{A2A Main Components}

An A2A agent is structured around several core components that define its behavior, capabilities, and interactions. These components serve as the operational and semantic building blocks for any agent to function within an agent-to-agent ecosystem.

 \textbf{Agent Card} acts as a self-description and discovery mechanism. It is a JSON-formatted document that publicly declares the agent’s metadata, including its name, version, description, supported skills, and authentication requirements. Client Agents rely on Agent Cards to discover and evaluate Remote Agents that can fulfill specific task criteria. As the primary entry point for coordination, an agent without an Agent Card is effectively invisible within the A2A system.

\textbf{Skills} represent the actionable capabilities offered by an agent. Each skill is described by a name, purpose, expected input parameters, and output format. Skills are invoked via Tasks and encapsulate the core utility the agent provides. An agent’s relevance and specialization are directly tied to the breadth and precision of its published skills.

\textbf{Task} is the atomic unit of work delegation. It specifies the skill to be executed, along with input parameters and contextual metadata. Tasks are issued by Client Agents and processed by Remote Agents, enabling asynchronous or synchronous collaboration. By structuring intention and invocation in a standardized format, Tasks allow A2A agents to operate interoperably across diverse systems.

\textbf{Messages} serve as the primary communication channel between agents. These encapsulate data exchange and coordination activities such as task submission, intermediate status updates, or artifact delivery—and can be composed of multiple typed parts including plain text, structured data, or file references. Without Messages, inter-agent interaction would not be possible, making them foundational to the A2A protocol.
 \textbf{Artifacts} are the tangible outputs of skill execution. Once a Remote Agent completes a task, it generates Artifacts that may contain structured responses, computed results, documents, or linked data. These outputs are transmitted back to the Client Agent, which may render them to the User or incorporate them into downstream processes. Artifacts represent the materialized knowledge or value created through agentic collaboration.

\subsection{A2A Transport Layer and Communication}

The A2A protocol supports multiple transport mechanisms to enable communication between Client and Remote Agents, tailored to support both synchronous and asynchronous workflows. When real-time streaming is required and supported by both parties, \textbf{Server-Sent Events (SSE)} can be employed. SSE establishes a persistent HTTP connection over which the Remote Agent can send live status updates or partial Artifacts to the Client Agent, facilitating continuous feedback during long-running tasks.

In scenarios where persistent connections are impractical such as mobile or distributed deployments—the protocol accommodates \textbf{Push Notifications}. These are implemented through a \texttt{PushNotificationService} interface that allows the Remote Agent to notify the Client about task progress or completion via out-of-band channels. This model is particularly suited for latency-tolerant workflows and background task orchestration.

All core task communications in A2A adhere to the \textbf{JSON-RPC 2.0} specification. This ensures a standardized format for method invocation, parameter passing, and result encapsulation. Additionally, Remote Agent discovery is bootstrapped via HTTP \texttt{GET} requests directed at the Agent’s endpoint, specifically retrieving its Agent Card as a structured representation of supported capabilities. Together, these mechanisms enable flexible, interoperable, and extensible communication across diverse runtime environments.

\subsection{A2A Remote Agent (Server) Lifecycle}

The lifecycle of a Remote Agent in the A2A protocol follows a structured progression through four key phases: \textbf{Creation}, \textbf{Operation}, \textbf{Update}, and \textbf{Termination}. Each phase reflects a distinct set of responsibilities critical to ensuring secure, discoverable, and reliable agent behavior.

\textbf{Creation} begins with the publication of the Agent Card, a JSON-formatted document served at \texttt{/.well-known/agent.json}, which declares metadata such as the agent’s name, version, supported skills, and authentication schemes. Once the Agent Card is made available, the agent service is deployed at a designated endpoint and configured to handle JSON-RPC 2.0 requests over HTTP. To complete the creation phase, the Remote Agent must implement the declared authentication mechanisms, enabling secure client verification and access control.

During \textbf{Operation}, the Remote Agent processes Tasks submitted by Client Agents in accordance with its advertised skills. This includes receiving structured Task payloads, executing the associated skill logic, and managing ongoing communication through JSON-RPC status messages and Artifact delivery. If supported, the agent may also stream asynchronous updates using Server-Sent Events (SSE) or deliver them through a registered \texttt{PushNotificationService}. The agent is responsible for maintaining internal task state throughout execution to ensure consistency and traceability across interactions.

In the \textbf{Update} phase, the Remote Agent refreshes its capabilities or configurations. This includes incrementing the version field in the Agent Card, adding new skills or authentication modes, and applying security patches to maintain compliance. The agent may also deprecate outdated features or legacy interfaces, ideally signaling these changes to clients through updated documentation or explicit lifecycle status fields.

Finally, during \textbf{Termination}, the Remote Agent gracefully winds down its operations. In-flight Tasks are driven to completion or transitioned to a terminal state, and any open SSE streams are closed. The service then deregisters its Agent Card—by removing or archiving the published metadata and releases allocated system resources, ensuring a clean shutdown with no residual exposure or stale endpoints. A well-defined Remote Agent lifecycle enhances discoverability, promotes interoperability, and ensures that agent-based collaboration remains secure, consistent, and predictable within the broader A2A ecosystem.

\subsection{Security Challenges and Mitigations Across the A2A Lifecycle}

A secure A2A deployment requires addressing threats across all lifecycle phases creation, operation, update, and termination. Table~\ref{tab:a2a-security-mitigations} consolidates principal vulnerabilities and corresponding mitigation strategies, drawn from official A2A protocol references and recent security analyses.

\begin{table}[ht]
   \small
  \centering
  \renewcommand{\arraystretch}{1.2}
  \rowcolors{2}{gray!10}{white}
  \captionsetup{justification=centering}
  \caption{A2A Lifecycle Security Challenges and Mitigation Strategies}
  \label{tab:a2a-security-mitigations}
\begin{tabularx}{\textwidth}{|
    >{\centering\arraybackslash}p{2.1cm}|
    >{\raggedright\arraybackslash}X|
    >{\raggedright\arraybackslash}X|
    >{\raggedright\arraybackslash}X|}

    \hline
    \rowcolor{gray!30}
    \textbf{Phase} & \textbf{Security Challenge} & \textbf{Threat Description} & \textbf{Mitigation Strategy} \\
    \hline

    \textbf{Creation}
    & Agent Card \& Manifest Spoofing
    & Adversaries may tamper with the Agent Card at \texttt{/.well-known/agent.json}, impersonating a trusted Remote Agent.
    & Digitally sign Agent Cards, verify checksums during retrieval, and harden CI/CD pipelines to prevent injection~\cite{google2024a2a}. \\

    \hline

    \textbf{Operation}
    & Task Injection \& Command Forgery
    & Malicious \texttt{tasks/send} or \texttt{tasks/sendSubscribe} calls may manipulate JSON-RPC to trigger unauthorized execution.
    & Enforce TLS, use JSON Web Signatures (JWS), validate schemas, and issue scoped capability tokens~\cite{redteaming2025}. \\

    & Push Notification Hijacking
    & Attackers may spoof SSE endpoints or intercept notifications, leading to fake updates or leakage.
    & Authenticate notification channels, isolate streams per session, and sign pushed events~\cite{firewalls2025}. \\

    \hline

    \textbf{Update}
    & Unauthorized Capability Injection \& Version Drift
    & Unauthorized actors may add hidden skills to Agent Cards, or clients may operate on outdated configurations.
    & Use immutable, versioned manifests, detect drift with GitOps, and require signed manifest diffs~\cite{mediumA2A2025}. \\

    \hline

    \textbf{Termination}
    & Orphaned Resources \& Audit Gaps
    & Tokens, SSE streams, or agent registrations may persist after use, complicating security audits.
    & Implement shutdown hooks, revoke credentials, and centralize audit logging with enforced retention~\cite{googleA2aBlog}. \\

    \hline
  \end{tabularx}
\end{table}




\begin{figure}
    \centering
    \includegraphics[width=\linewidth]{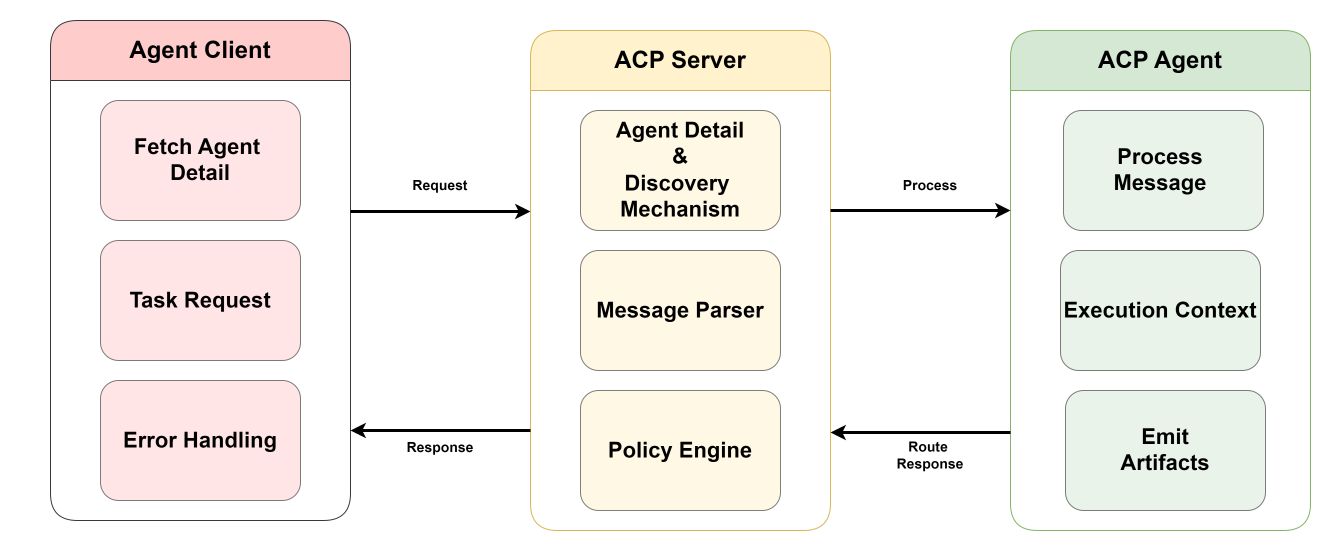}
    \caption{An overview of ACP}
    \label{fig:ACP}
\end{figure}



\section{ACP Architecture}

The Agent Communication Protocol (ACP) architecture is a modular, HTTP-based system designed to facilitate interaction between clients and agents. It is composed of three primary components: the ACP Client, the ACP Server, and one or more ACP Agents. Clients communicate with servers using standardized REST APIs, sending structured multi-modal messages that agents process and responds.

The \textbf{ACP Client} initiates communication by submitting requests in ACP-compliant format. It supports message composition using ordered message parts, session-based interactions for multi-turn workflows, and both synchronous and streaming execution modes. Error responses follow a unified structure, and standard HTTP authentication methods such as Bearer tokens, Basic Auth, and JWTs are supported.

The \textbf{ACP Server} acts as a middleware component, translating external HTTP requests into internal agent executions. It manages agent lifecycle phases including execution and optional session persistence. The server is stateless by default, making it compatible with load balancers and orchestration platforms such as Kubernetes. Secure communication is achieved via TLS, and observability is supported through OpenTelemetry-based tracing and metrics.

The \textbf{ACP Agent} is the execution component defined using a decorator-based configuration. Agents process structured requests consisting of ordered message parts and generate responses that conform to ACP’s message format. They support both stateless and session-aware operation, including features such as await/resume for interactive use cases. Agent metadata enables discovery through runtime APIs, offline packaging, or public manifest files.

Together, these three components form a framework that supports agent communication through standardized protocols, structured messaging, and flexible deployment strategies.

\subsection{ACP Main Components}

ACP interactions are governed by a set of core components that define agent behavior, enable runtime interoperability, and standardize task communication. Central to this architecture is the \textbf{Agent Detail}, a self-descriptive JSON or YAML document that serves as the agent’s public identity and capability profile. It provides essential metadata, including the agent’s name, available operations, supported content types, authentication schemes, and runtime diagnostics. Clients rely on Agent Detail as a precondition for invocation, enabling trust and selection without bespoke integration.

Complementing this, \textbf{Discovery Mechanisms} allow clients to locate agents dynamically at runtime. These mechanisms may be centralized—such as registry APIs—or decentralized, including manifest files hosted under well-known URLs (e.g., \texttt{/.well-known/agent.yml}) or embedded within deployment metadata like container labels. This discoverability layer decouples client logic from fixed configurations and supports scalable agent networks.

Once an agent is located, clients issue a \textbf{Task Request}, a structured unit of delegated work. Task Requests are composed of ordered message parts that specify the target operation and include textual inputs, binary payloads, or references to externally hosted data. This design accommodates both synchronous calls and long-running, asynchronous tasks.

All requests and responses conform to ACP’s \textbf{Message Structure}, which standardizes the communication envelope. Each message is an ordered list of parts, with explicit MIME \texttt{content\_type} annotations and either embedded \texttt{content} or dereferenceable \texttt{content\_url} values. Optional \texttt{name} attributes enable the use of semantically tagged Artifacts, facilitating downstream interpretation.

Finally, the result of agent execution is encapsulated in one or more \textbf{Artifacts}. These may consist of structured JSON outputs, plain text completions, binary files, or even nested message references. Artifacts are delivered as part of the Message Structure response and are subsequently rendered, stored, or chained into additional agent workflows, ensuring extensibility and composability across ACP-enabled systems.

\subsection{ACP Agent Lifecycle}

The lifecycle of an ACP agent closely mirrors the A2A framework’s four canonical phases: \textit{Creation}, \textit{Operation}, \textit{Update}, and \textit{Termination}. Each stage ensures that agent behavior remains discoverable, interoperable, and secure throughout its active deployment.

\textbf{Creation} begins with the configuration and deployment of the agent. This involves declaring the agent’s capabilities and metadata through an Agent Detail manifest, which is made accessible via an ACP-compliant server, such as an ASGI-based or built-in implementation. The agent is initialized with authentication mechanisms and routing logic that collectively secure both service discovery and downstream task execution.

During \textbf{Operation}, agents process structured \texttt{sendTask} requests submitted by clients. These requests contain encoded parameters required for task execution. The ACP runtime supports synchronous execution as well as incremental streaming of intermediate results. Each task progresses through well-defined states—such as \texttt{created}, \texttt{in\_progress}, or \texttt{awaiting} that are managed by the ACP execution engine. For multi-turn workflows, session-level persistence ensures continuity of context across multiple interactions.

In the \textbf{Update} phase, the Agent Detail manifest is refreshed to reflect changes in the agent’s behavior or capabilities. These updates may include new operations, supported MIME types, or version increments. Importantly, the discovery process is resilient to such changes: clients querying the agent registry retrieve the latest manifest without requiring direct API modification, thereby preserving backward compatibility.

\textbf{Termination} involves the graceful decommissioning of the agent. All active tasks are driven to completion, ongoing streams are closed, and the agent’s manifest is deregistered or marked as inactive to prevent future discovery. Any allocated resources are released, and session data is finalized to ensure a clean and auditable shutdown.

\subsection{Security Considerations Across the ACP Lifecycle}

ACP-based systems face distinct security challenges as they progress through their lifecycle phases—from agent registration to shutdown. Table~\ref{tab:acp-security-mitigations} summarizes key threats and their corresponding mitigation strategies, grounded in recent red-teaming, protocol, and platform-level research.

\begin{table}[ht]
  \small
  \centering
  \renewcommand{\arraystretch}{1.2}
  \rowcolors{2}{gray!10}{white}
  \captionsetup{justification=centering}
  \caption{ACP Lifecycle Security Challenges and Mitigation Strategies}
  \label{tab:acp-security-mitigations}
\begin{tabularx}{\textwidth}{|
    >{\centering\arraybackslash}p{2.1cm}|
    >{\raggedright\arraybackslash}X|
    >{\raggedright\arraybackslash}X|
    >{\raggedright\arraybackslash}X
|}
    \hline
    \rowcolor{gray!30}
    \textbf{Phase} & \textbf{Security Challenge} & \textbf{Threat Description} & \textbf{Mitigation Strategy} \\
    \hline

    \textbf{Creation}
    & Metadata Spoofing \& Supply Chain Attacks
    & Attackers may publish forged Agent Detail manifests (e.g., \texttt{/.well-known/agent.yml}) to impersonate agents or inject malicious skills.
    & Digitally sign all manifests, verify at discovery, and enforce CI/CD signature checks and artifact validation~\cite{beeai2024acp}. \\

    \hline

    \textbf{Operation}
    & Message Tampering \& MITM
    & Adversaries can intercept or alter \texttt{sendTask} or \texttt{getTask} RPC calls, leading to payload injection or message corruption.
    & Use TLS for transport security and sign each message part with JWS~\cite{redteaming2025}. \\

    & Auth Flaws \& Unauthorized Access
    & Weak bearer token enforcement may allow unauthorized execution or task disruption.
    & Apply capability-scoped, short-lived tokens; enforce mutual TLS with identity revocation~\cite{mediumA2A2025}. \\

    \hline

    \textbf{Persistence}
    & Session Hijacking \& Privacy Leaks
    & Replay attacks or token theft may occur in long-lived sessions without proper binding or encryption.
    & Rotate session IDs, encrypt persisted context, and minimize token lifetimes~\cite{firewalls2025}. \\

    \hline

    \textbf{Update}
    & Version Rollback \& Config Drift
    & Stale manifests or software may reintroduce patched vulnerabilities post-update.
    & Enforce immutable, versioned manifests and use GitOps to detect drift~\cite{acmAgents2025}. \\

    \hline

    \textbf{Termination}
    & Orphaned Resources \& Audit Gaps
    & Failing to revoke tokens or close SSE streams complicates cleanup and forensics.
    & Drain active tasks, revoke credentials, and centralize audit logging with retention policies~\cite{acmAgents2025}. \\

    \hline
  \end{tabularx}
\end{table}

\section{ANP Architecture}
Agent Network Protocol (ANP) is a decentralized, peer-to-peer communication standard designed for cross-platform agent interoperability on the open internet. ANP enables agents to autonomously discover, authenticate, and interact using structured metadata and AI-native data exchange. The following sections align ANP's architecture with a standardized lifecycle and modular framework, modeled consistently with MCP, A2A, and ACP.

\begin{figure}
    \centering
    \includegraphics[width=0.55\linewidth]{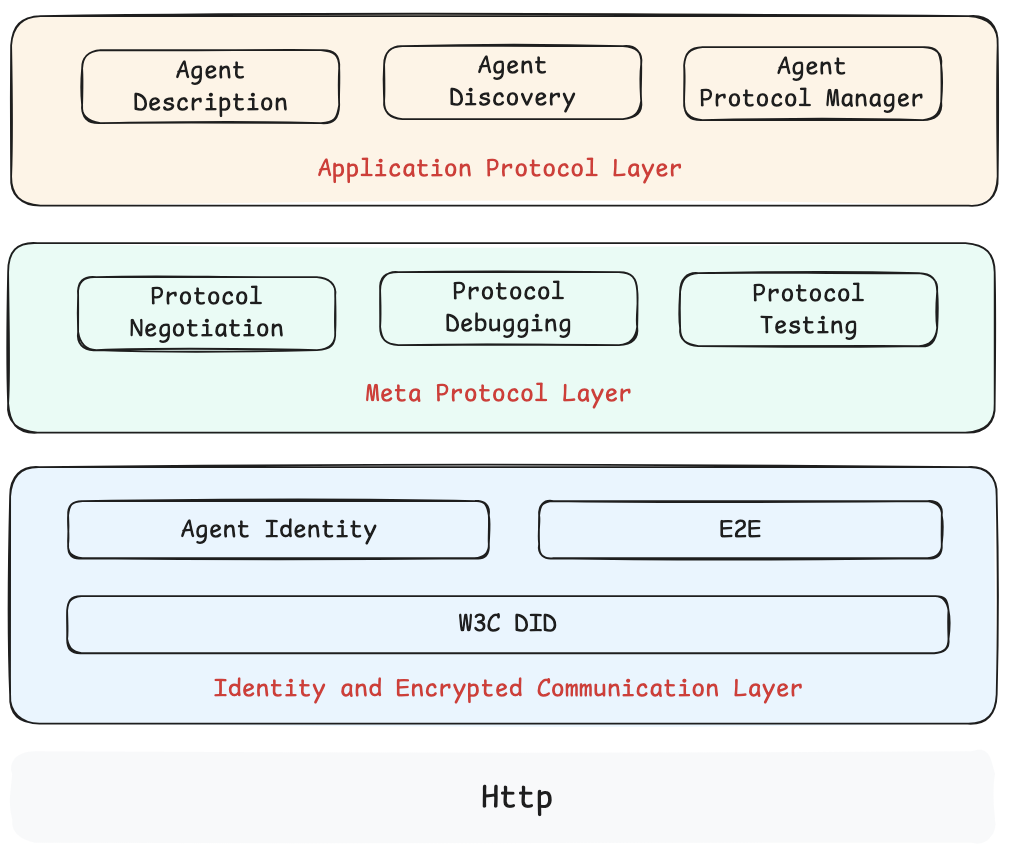}
    \caption{An overview of ANP \cite{anp2024website,anp2024github}}
    \label{fig:ANP}
\end{figure}

\subsection{Core Components}
The Agent Network Protocol (ANP) is underpinned by a set of foundational components that collectively support decentralized identity, semantic self-description, discovery, and adaptive interaction. At the core is the \textbf{Agent Identity}, which employs Decentralized Identifiers (DIDs) to uniquely identify agents across platforms. Specifically, ANP adopts the \texttt{did:wba} method, where each identifier corresponds to an HTTPS-hosted DID document, thereby leveraging existing Web infrastructure for decentralized identity resolution.

Building upon this identity layer is the \textbf{Agent Description}, implemented through the Agent Description Protocol (ADP). These JSON-LD formatted documents contain structured metadata about the agent, including its name, capabilities, supported protocols, authentication schemes, and service endpoints. They serve as the agent’s publicly accessible profile, facilitating interoperability and semantic understanding.

Agents expose their presence and capabilities through a \textbf{Discovery Directory}, typically located at the standardized \texttt{.well-known/agent-descriptions} endpoint. This directory enables both human users and automated systems to retrieve a list of available agents under a given domain, forming the basis for scalable agent indexing and search.

To support interaction, ANP accommodates two categories of communication interfaces: \textbf{Structured Interfaces}, such as JSON-RPC and OpenAPI, and \textbf{Natural Language Interfaces}, defined via YAML or equivalent schema files. Both interface types are declared within the agent’s description and enable flexible interaction patterns suited to varying complexity and use cases.

Finally, the \textbf{Meta-Protocol Negotiator} facilitates dynamic protocol alignment between agents. This mechanism allows agents to exchange natural language descriptions of their communication requirements and capabilities, from which compatible interaction protocols can be negotiated and instantiated. By supporting runtime adaptability and negotiation, this layer ensures seamless interoperability even among heterogeneous agent ecosystems.

\subsection{ANP Agent Lifecycle}

The ANP agent lifecycle adheres to the canonical phases of \textit{Creation}, \textit{Operation}, \textit{Update}, and \textit{Termination}, reflecting the decentralized design principles of the Agent Network Protocol. Each phase ensures that agents remain discoverable, verifiable, and interoperable within a globally distributed agent ecosystem.

\textbf{Creation} initiates with the generation of a decentralized identifier (DID) using the \texttt{did:wba} method. This identifier is associated with a publicly resolvable HTTPS endpoint hosting the agent’s DID document. In parallel, the agent prepares a self-descriptive Agent Description (ADP) document in JSON-LD format, detailing its services, supported protocols, and authentication mechanisms. The ADP is then published under a standardized path such as \texttt{/.well-known/agent-descriptions}, enabling web-based discovery or optional registration with search agents.

During the \textbf{Operation} phase, agents authenticate and interact via cryptographic credentials defined in their DID documents. All communications follow structured interaction models declared in the ADP—such as JSON-RPC for precise invocation or YAML-based interfaces for natural language negotiation. Secure transport is established using HTTPS and, where applicable, real-time communication is supported through mechanisms such as Server-Sent Events (SSE) or long polling. Agents act autonomously or cooperatively by invoking external services, interpreting requests, and returning results in a standardized format.

The \textbf{Update} phase allows agents to revise their ADP documents and associated DID metadata to reflect evolving capabilities or interface changes. These updates are automatically propagated through recurring crawls by indexing services or explicitly refreshed via active discovery endpoints. Because agent identity and service descriptions are independently versioned and published, clients can dynamically adapt to updates without breaking existing integrations.

\textbf{Termination} involves the intentional deactivation of an agent. This includes the removal or archival of its DID document and the depublication of its ADP endpoint from discovery directories. Any issued authentication tokens, access credentials, or associated metadata must be revoked to ensure security. A clean shutdown preserves the integrity of the agent ecosystem by preventing stale or orphaned entries from persisting in discovery indexes or trusted registries.

\subsection{Transport and Format}
ANP relies on HTTP(S) for transport and JSON-LD for data formatting. Schema.org vocabularies and contexts like `ad:` are used for semantic clarity. Structured interfaces such as JSON-RPC and OpenAPI are compatible and embedded via ADP.

\subsection{Security Considerations Across the ANP Lifecycle}
Table~\ref{tab:anp-security-mitigations} summarizes major threats and corresponding mitigations across the ANP lifecycle.

\begin{table}
    \small
  \centering
  \renewcommand{\arraystretch}{1.3}
  \rowcolors{2}{gray!10}{white}
  \captionsetup{justification=centering}
  \caption{ANP Lifecycle Security Challenges and Mitigation Strategies}
  \label{tab:anp-security-mitigations}
  \begin{tabularx}{\textwidth}{|
      >{\centering\arraybackslash}p{2.3cm} |
      >{\raggedright\arraybackslash}X |
      >{\raggedright\arraybackslash}X |
      >{\raggedright\arraybackslash}X |
  }
    \hline
    \rowcolor{gray!30}
    \textbf{Phase} & \textbf{Security Challenge} & \textbf{Threat Description} & \textbf{Mitigation Strategy} \\
    \hline

    \textbf{Creation}
    & Identity Spoofing
    & DID documents may be spoofed or hosted insecurely, leading to agent misidentification.
    & Enforce HTTPS-hosted DIDs, verify with DNS records, and require DID signature validation. \\

    \textbf{Operation}
    & Unverified Agents
    & Malicious actors may bypass DID checks or use spoofed credentials.
    & Authenticate via DID public keys and validate humanAuthorization for sensitive actions. \\

    & Interface Tampering
    & Agents may alter structured interfaces or inject into natural language endpoints.
    & Require cryptographic signing of interfaces and log access events with source metadata. \\

    \textbf{Update}
    & Stale Descriptions
    & Outdated or manipulated agent metadata may deceive clients.
    & Automate crawling of agent descriptions and validate against known-good hashes. \\

    \textbf{Termination}
    & Orphaned Identifiers
    & Expired DIDs or agent declarations (ADPs) may persist in registries or caches.
    & Use expiration timestamps and require revocation signaling during deregistration. \\

    \hline
  \end{tabularx}
\end{table}

\begin{table}[h]
    \small
  \centering
  \renewcommand{\arraystretch}{1.2}
  \rowcolors{2}{gray!10}{white}
  \caption{Comparison of MCP, ACP, A2A, and ANP Protocols}
  \captionsetup{justification=centering}
  \label{tab:agent-protocol-comparison}
  \begin{tabularx}{\textwidth}{|
      >{\raggedright\arraybackslash}p{2.2cm}|
      >{\raggedright\arraybackslash}X|
      >{\raggedright\arraybackslash}X|
      >{\raggedright\arraybackslash}X|
      >{\raggedright\arraybackslash}X|
  }
    \toprule
    \rowcolor{gray!25}
    \textbf{Aspect} &
    \textbf{MCP (Model Context Protocol)} &
    \textbf{ACP (Agent Communication Protocol)} &
    \textbf{A2A (Agent-to-Agent Protocol)} &
    \textbf{ANP (Agent Network Protocol)} \\
    \midrule

    \textbf{Architecture Model} &
    Client–Server with JSON-RPC primitives &
    Brokered Client–Server (Registry + Task Routing) &
     Peer-like Client $\leftrightarrow$ Remote Agent
&
    Decentralized Peer-to-Peer \\

    \textbf{Agent Discovery} &
    Manual registration or static URL lookup &
    Registry-based   &
    Agent Card retrieval via HTTP &
    Search Engine Discovery \\

    \textbf{Identity \& Auth} &
    Token-based auth; supports DIDs optionally &
    Bearer tokens, mutual TLS, JWS &
    DID-based handshake or out-of-band headers &
    Decentralized Identifiers (DID), especially \texttt{did:wba} \\

    \textbf{Message Format} &
    JSON-RPC 2.0 with Prompts, Tools, Resources &
    Structured multipart messages with MIME-typed parts &
    Task + Artifact messaging over JSON &
    JSON-LD with Schema.org and ADP/Meta-Protocol negotiation \\

    \textbf{Core Components} &
    Tools, Prompts, Resources, Sampling &
    Agent Detail, Message, Task Request, Artifact &
    Agent Card, Task, Message, Artifact &
    DID Document, Agent Description, Meta-Protocol, Structured Interface \\

    \textbf{Transport Layer} &
    HTTP, Stdio, Server-Sent Events (SSE) &
    HTTP with incremental streams &
    HTTP with optional SSE + Push Notifications &
    HTTP with JSON-LD over TLS \\

    \textbf{Session Support} &
    Stateless + optional persistent tool context &
    Session-aware with run state tracking &
    Session-aware or stateless; client-managed IDs &
    Stateless; DID-authenticated tokens used across connections \\

    \textbf{Target Scope} &
    LLM $\leftrightarrow$ External Tool/Service integration &
    Model-Agnostic, Infrastructure-level agents &
    Trusted enterprise task delegation &
    Open Internet agent interconnectivity \\

    \textbf{Primary Use Case} &
    Augment LLMs with external capabilities (e.g., code, search) &
    Secure, typed message exchange for diverse agents &
    Multi-agent workflows within organizational trust boundaries &
    Cross-platform agent discovery, secure P2P execution \\

    \textbf{Strengths} &
    Tight LLM integration; resource injection &
    Multimodal messaging; brokered registry; tool modularity; Offline Agent Discovery &
    Inter-agent negotiation; artifact-driven delegation &
    DID-based trustless identity; AI-native protocol negotiation \\

    \textbf{Limitations} &
    Centralized server assumption; prompt injection risks &
    Registry required; strong assumptions on server control &
    Enterprise-centric; assumes agent catalog &
    High negotiation overhead; evolving adoption ecosystem \\

    \bottomrule
  \end{tabularx}
\end{table}

\section{Comparison of Agent Protocols}

To facilitate a clearer understanding of how major agent interoperability protocols differ, Table~\ref{tab:agent-protocol-comparison} presents a side-by-side comparison of four widely discussed frameworks: Model Context Protocol (MCP), Agent Communication Protocol (ACP), Agent-to-Agent Protocol (A2A), and Agent Network Protocol (ANP). This structured analysis highlights their architectural choices, messaging formats, discovery methods, session models, and intended use cases, offering insights into their suitability across diverse deployment scenarios.

\section{Phased Adoption Roadmap for Agent Interoperability}

This section presents a practical, multi-stage deployment strategy for agent interoperability based on protocol maturity, integration complexity, and use case alignment. The roadmap enables organizations to adopt suitable agent communication standards while supporting scalability, composability, and security.

\subsection{Stage 1 – MCP for Tool Invocation}
The initial phase involves adopting the Model Context Protocol (MCP) to enable structured and secure interaction between large language models (LLMs) and external tools or resources. MCP operates over a JSON-RPC-based client-server model and is well-suited for use cases focused on tool invocation, deterministic execution, and typed input/output. This stage establishes a foundation for context enrichment in single-model systems.

\subsection{Stage 2 – ACP for Agent Communication}
The second phase introduces the Agent Communication Protocol (ACP), a general-purpose, REST-native protocol designed for asynchronous and synchronous communication between independent agents. ACP supports MIME-typed multipart messages, ordered message parts, and streaming. These capabilities enable standardized messaging across agents built on different frameworks, facilitating collaboration and interoperability across organizational and technological boundaries.

\subsection{Stage 3 – A2A for Enterprise Collaboration}
In enterprise-oriented deployments, the Agent-to-Agent (A2A) protocol facilitates multi-agent interaction through Agent Cards and artifact exchanges. It supports dynamic discovery and structured messaging using predefined types, making it well-suited for coordinated workflows among stateless or stateful agents within trusted organizational contexts.

\subsection{Stage 4 – ANP for Open Agent Markets}
The final phase involves extending interoperability to the open internet using the Agent Network Protocol (ANP). ANP facilitates decentralized agent discovery, DID-based identity verification, and peer-to-peer communication using JSON-LD graphs. It provides the foundation for scalable, cross-platform agent marketplaces and AI-native web interaction. This phased approach enables organizations to adopt agent communication protocols progressively, maximizing interoperability while minimizing integration complexity at each stage.

\section{Conclusion}

As autonomous agents powered by large language models proliferate across domains, the demand for secure, modular, and interoperable communication grows increasingly urgent. This survey presented a structured analysis of four emerging protocols MCP, ACP, A2A, and ANP that each address distinct layers of agent interoperability. By unifying tool invocation, multimodal messaging, task coordination, and decentralized discovery, these protocols collectively form the foundation for scalable multi-agent systems. The comparative evaluation demonstrates that no single protocol suffices across all contexts; instead, a phased, complementary adoption strategy—beginning with MCP and progressing through ACP and A2A to ANP offers a practical pathway for deploying agent ecosystems. Future research should explore protocol interoperability bridges, trust frameworks for agent collaboration, and standardized evaluation benchmarks to accelerate adoption and ensure resilience in real-world deployments. These foundational efforts will be critical for advancing the next generation of intelligent, networked agents.

\bibliographystyle{unsrt}  
\bibliography{references}

\end{document}